\documentclass[10pt,twocolumn,letterpaper]{article}

\usepackage{wacv}              

\definecolor{wacvblue}{rgb}{0.21,0.49,0.74}
\usepackage[pagebackref,breaklinks,colorlinks,allcolors=wacvblue]{hyperref}

\usepackage{threeparttable}
\usepackage{amsmath}
\usepackage{amssymb}
\usepackage{bm}
\usepackage{multirow}
\usepackage{pifont}
\usepackage{array}
\usepackage{tabularx}
\usepackage{adjustbox}
\usepackage{makecell}   
\usepackage[ruled,vlined]{algorithm2e}

\usepackage{array}
\usepackage{adjustbox}
\usepackage{tabularx}
\newcolumntype{L}[1]{>{\raggedright\arraybackslash}p{#1}}
\def\wacvPaperID{2354} 
\def\confName{WACV}
\def\confYear{2027}

\title{BEV3SR: Evidence-Aware 3D Spatial Reasoning from Video}

\author{Fan Zhu, \quad Rui Yu\\
University of Louisville\\
Louisville, KY, USA\\
{\tt\small \{fan.zhu, rui.yu\}@louisville.edu}}

\begin{document}
\maketitle
\raggedbottom
\begin{abstract}
        3D spatial reasoning from video remains challenging for VLMs because videos provide sequential, view-local observations rather than persistent metric scene records. Many questions must be answered in a query-specified perspective rather than the observed camera viewpoint, and prompting with a fixed set of RGB views may miss sparse or occluded visual cues. To address these gaps, we introduce \textbf{BEV3SR}, a training-free and backbone-agnostic framework for \mbox{RGB-D} videos with camera parameters. BEV3SR turns video observations into queryable metric bird’s-eye-view (BEV) scene memory and formalizes each query into an evidence-aware spatial task. The \textbf{Spatial Task Formalizer} specifies the query’s reference frame, spatial constraints, objective, and evidence requirement. The evidence requirement determines the appropriate reasoning pathway. A query treated as requiring geometric evidence is handled by the \textbf{Geometric Reasoner}, which deterministically computes the answer using the metric BEV records in the query-specified reference frame. Otherwise, a query treated as requiring visual evidence is passed to the \textbf{Visual Evidence Aligner}, which uses BEV-indexed object records and their associated supporting views to retrieve object-consistent and scene-context RGB frames for VLM grounding. Within a standard 16-frame budget, BEV3SR consistently improves different VLM backbones across multiple 3D scene QA benchmarks, delivering large gains on spatially intensive questions and significant improvements on open-ended scene QA. These results demonstrate that BEV3SR offers an effective, efficient, and interpretable framework for explicit 3D spatial reasoning from video.
\end{abstract}
    
\section{Introduction}
\label{sec:intro}
With the rapid progress of multimodal large language models (MLLMs) \cite{achiam2023gpt,comanici2025gemini,liu2024deepseek,li2023blip2,liu2023llava,wang2025internvl3,bai2025qwen3}, visual-spatial understanding is expanding beyond local spatial judgments in isolated images toward scene-level reasoning \cite{azuma2022scanqa,ma2022sqa3d,yang2025thinking} about the 3D physical environments. The video approximates how humans and embodied agents observe physical environments through a moving viewpoint, provides rich multi-view evidence about object appearances, scene layouts, and spatial relations, and has therefore become a natural setting for advancing spatial reasoning in 3D scenes \cite{yang2025thinking, lin2025mmsi,majumdar2023openeqa,zhang2026flatland, li2025spatialladder}. However, recent studies suggest that current MLLMs still struggle to leverage video observations for reliable 3D spatial reasoning. This limitation reflects several challenges repeatedly identified in prior work.

\begin{figure*}[!t]
    \centering
    \includegraphics[width=0.8\textwidth]{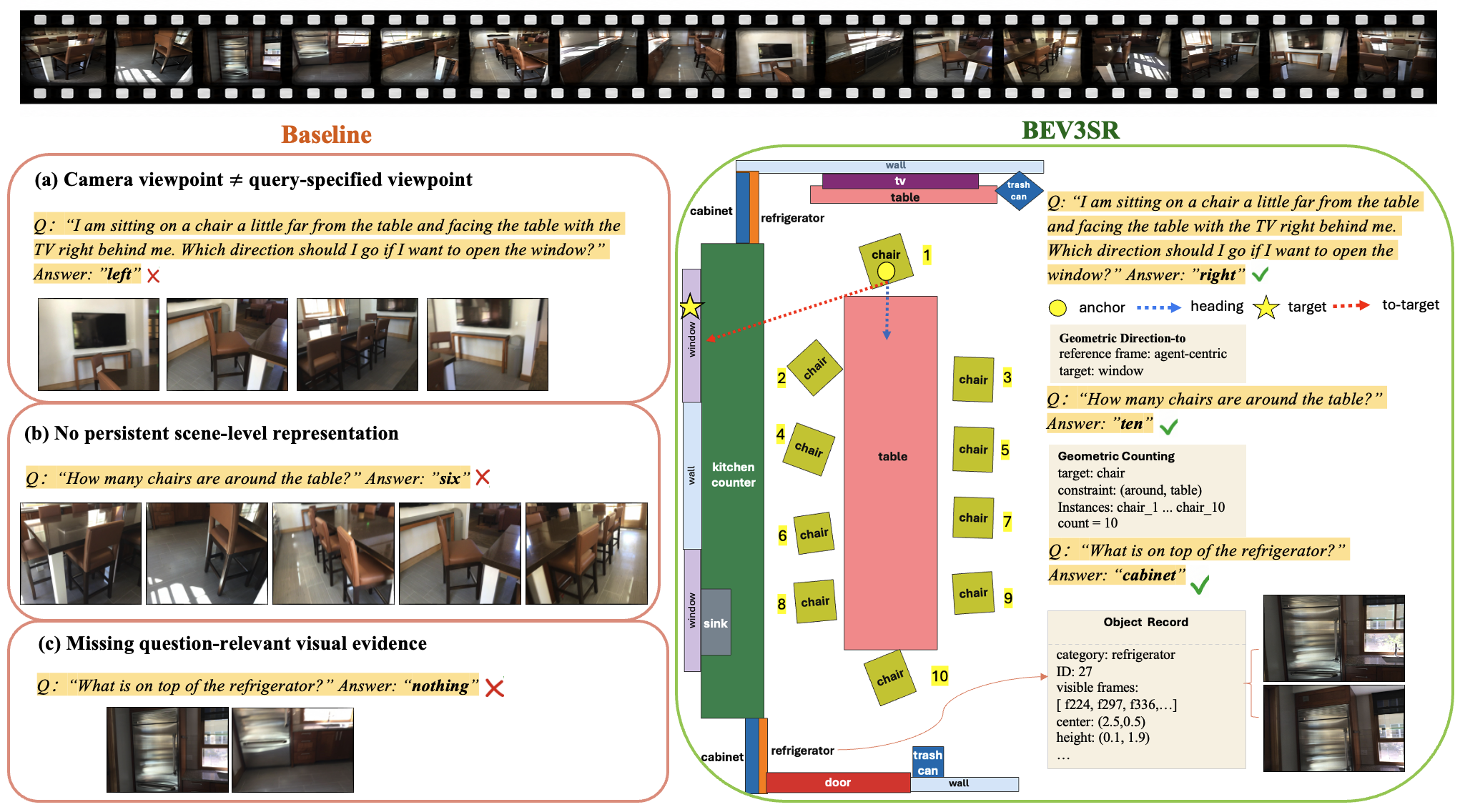}
\vspace{-2mm}
    \caption{Three limitations of the VLM baseline and how the proposed
BEV3SR addresses them. The baseline uses Qwen3-VL-8B~\cite{bai2025qwen3} prompted with 16 uniformly sampled RGB frames.}
\vspace{-2mm}
    \label{fig:limitations}
\end{figure*}

First, video-based reasoning requires models to consolidate fragmented, camera-centric observations into a persistent metric scene representation. However, existing Vision Language Models (VLMs) often rely on local or frame-level visual evidence, making it difficult to maintain a unified scene-level context across multiple views \cite{yang2025thinking, chen2026egomind, xu2026multi}. 
Second, many 3D scene questions must be answered in a query-specified allocentric reference frame, rather than from the egocentric viewpoint of camera. Existing MLLMs often conflate these two coordinate systems: they may recognize the relevant objects but still answer according to the camera view, mistaking image-left or image-right for the agent’s left or right \cite{wang2026allocentric,zhang2025do,zhang2025spinbench}. 
Third, finite-budget video prompting does not guarantee question-relevant evidence retrieval. Because MLLMs can only ingest a limited number of sampled frames, relevant objects or spatial relations may be missed when they are sparse, occluded, briefly visible, or visually verifiable only from particular viewpoints. Thus, even a strong VLM may answer incorrectly because the necessary visual evidence is not present in its input context \cite{zhang2026revsi, zhao2026cov}. We illustrate these three limitations in Fig.~\ref{fig:limitations} using a representative 16-frame uniformly sampled prompting VLM baseline~\cite{bai2025qwen3}.

To address these limitations, recent studies have explored two broad directions for enhancing the 3D awareness of MLLMs. The first direction improves implicit 3D representations inside MLLMs by injecting explicit 3D signals \cite{chen2024grounded3dllm,huang2024chat,zhu2025llava,GPT4Scene,zheng2024video3dllmlearningpositionaware}, by aligning MLLMs with 3D reconstruction or geometric priors from feed-forward geometry foundation models \cite{wu2026spatial,fan2026vlm,zhang2026spatialstack,zhang2026make}, or by introducing spatially supervised training objectives \cite{xu2026multi,hu2026g,qu2026loc3r}. The second direction makes spatial reasoning more explicit at inference time through active or synthetic viewpoint exploration \cite{zhao2026cov,yang2026mindjourney}, chain-of-thought \cite{zhang2026spatial}, or geometric tools \cite{chen2026geometrically}. These methods either enrich the model’s implicit 3D scene representation, acquire additional query-relevant views at test time, or make reference-frame reasoning and metric computation more explicit. However, they typically address these limitations separately rather than providing a scene memory that jointly supports query-specified geometric reasoning and finite-budget visual evidence alignment.

Motivated by these observations, we revisit BEV not as a conventional top-down image or a visual prompt, but as a BEV-indexed metric scene memory for 3D scene question answering. Its key role is organization rather than visualization: it consolidates multi-frame observations into persistent object-level records, where each object is linked to a stable metric location, spatial extent, and supporting RGB views in a shared scene-level coordinate system. This allows BEV to serve as a queryable spatial interface, supporting both view evidence retrieval for visual grounding and deterministic metric computation for geometric reasoning.

Based on this design, we propose \textbf{BEV3SR}, a training-free and backbone-agnostic framework for \underline{\textbf{BEV}}-indexed metric scene memory-based \underline{\textbf{3}}D \underline{\textbf{S}}patial \underline{\textbf{R}}easoning.
BEV3SR consists of three coupled components. First, it turns \mbox{RGB-D} video with camera parameters into queryable metric BEV scene memory with object-level records and supporting views. Second, the \textbf{Spatial Task Formalizer} converts each natural-language query into an evidence-aware spatial task, specifying the query reference frame, spatial constraints, objective, and evidence requirement. 
Third, the evidence requirement determines the appropriate reasoning pathway. This design is conceptually aligned with the dual-stream view of visual cognition \cite{goodale1992separate}, which separates object-oriented perception from spatial reasoning for action. A query treated as requiring geometric evidence is handled by the \textbf{Geometric Reasoner}, which deterministically computes the answer using metric BEV records in the query-specified reference frame. Otherwise, a query treated as requiring visual evidence is passed to the \textbf{Visual Evidence Aligner}, which uses BEV-indexed object records and associated supporting views to retrieve object-consistent and scene-context RGB frames for VLM grounding. In this way, BEV3SR unifies persistent scene memory, finite-budget visual evidence alignment, and geometric reasoning within a single 3D scene QA framework.

Our contributions are threefold. First, we formulate video-based 3D spatial reasoning as evidence-aware task execution over BEV-indexed metric scene memory, positioning BEV as an object-level spatial index rather than a visual prompt. Second, we introduce explicit spatial task formalization for natural-language 3D scene QA, enabling query-defined reference-frame reasoning and evidence-based pathway selection. Third, we instantiate this formulation with two complementary modules for deterministic geometric reasoning and finite-budget visual evidence alignment, and demonstrate consistent gains over baseline method across multiple VLM backbones. Under the same 16-frame budget, BEV3SR improves SQA3D from 51.6/55.0 to 62.6/64.8 EM/EM-R and ScanQA from 27.8 to 30.6 EM and 82.6 to 95.6 CIDEr. On the static spatial subset of VSI-Bench, BEV3SR achieves 63.8 Avg., outperforming the baseline by 10.0 points and remaining competitive with recent 3D-aware models, all without task-specific training.

\section{Related Work}
\label{sec:relatedworks}
\paragraph{3D-aware MLLMs for spatial reasoning.}
Recent work improves the 3D awareness of 2D-pretrained MLLMs by providing explicit spatial signals or spatial supervision. One line injects 3D structure into the input or latent representation, including point-cloud features, object-level 3D tokens, depth-backprojected coordinates, camera-pose encodings, and BEV/object-marker cues for global-local grounding \cite{3dllm, chen2024ll3da, zhu2024unifying, fu2024scenellmextendinglanguagemodel,chen2024grounded3dllm, huang2024chat, huang2024embodied,zhu2025llava,GPT4Scene,zheng2024video3dllmlearningpositionaware}. Recent methods further incorporate reconstruction-aware priors from feed-forward geometry foundation models or introduce spatial supervision for layout, pose, correspondence, and situation modeling \cite{wu2026spatial,fan2026vlm,zhang2026spatialstack,zhang2026make,xu2026multi,hu2026g,qu2026loc3r}. These approaches demonstrate the value of 3D structure, but spatial information is still primarily absorbed into model representations and the reasoning process remains largely implicit. In contrast, BEV3SR maintains an explicit metric scene memory outside the VLM, making object records, reference frames, and supporting evidence queryable and controllable.

\paragraph{Explicit spatial reasoning at inference time.}
Another line makes spatial reasoning more explicit during inference. Active or synthetic viewpoint methods acquire additional observations by selecting, moving, or synthesizing views according to the query \cite{zhao2026cov,yang2026mindjourney}. Structured prompting and chain-of-thought methods decompose spatial questions into intermediate scene descriptions or cognitive map, coordinates, or reasoning steps \cite{yang2025thinking,zhang2026spatial}. Tool-augmented methods offload geometric computation to external modules \cite{chen2026geometrically, han2025tiger}. These methods improve interpretability by reducing purely implicit reasoning. BEV3SR is complementary but targets a different setting: video-based 3D scene QA under a fixed visual budget. Instead of repeatedly acquiring new views or applying tools mainly to already provided views, or task-specified regions, BEV3SR treats finite-budget evidence selection itself as part of the reasoning problem. It uses a reusable BEV-indexed metric memory to support both query-specified geometric reasoning and visual evidence alignment for perception-dependent answers.

\section{Methodology}
\begin{figure*}[!t]
    \centering
    \includegraphics[width=0.78\textwidth]{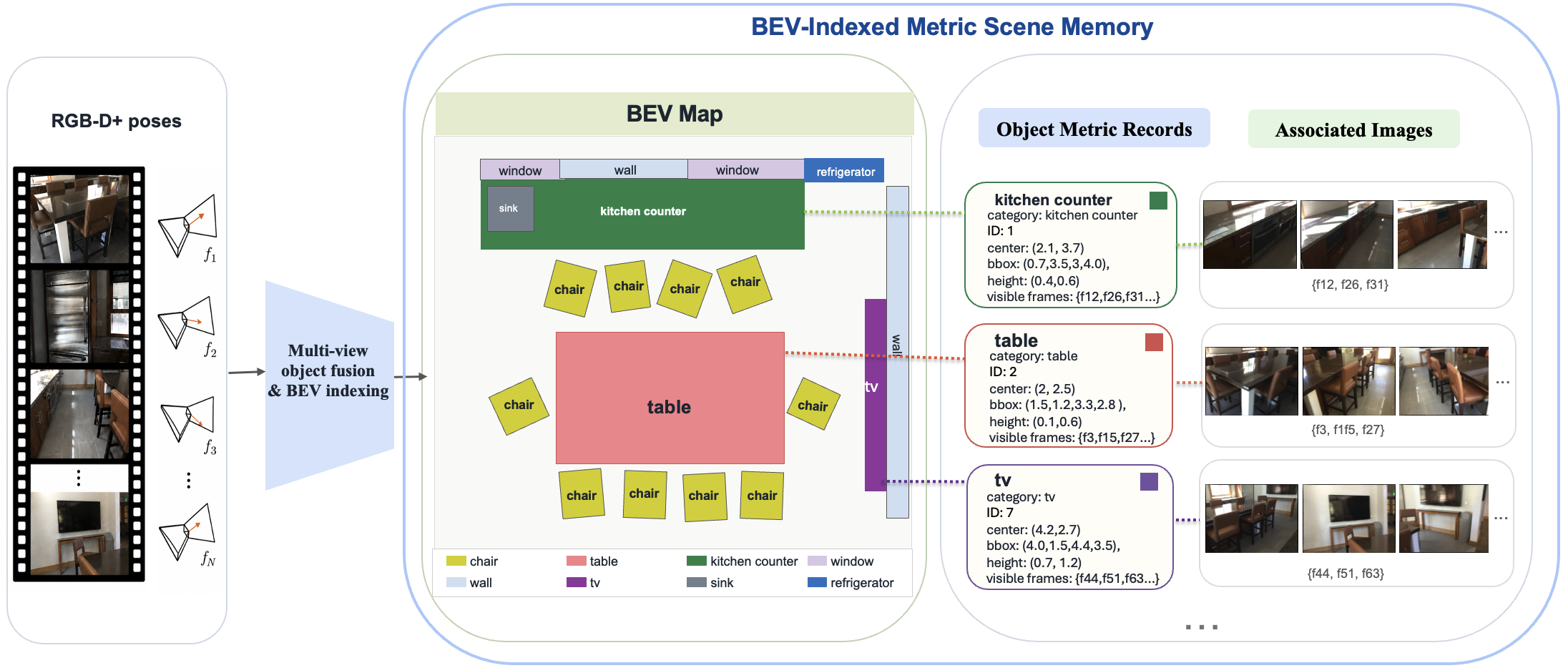}
    \vspace{-0.6em}
    \includegraphics[width=0.78\textwidth]{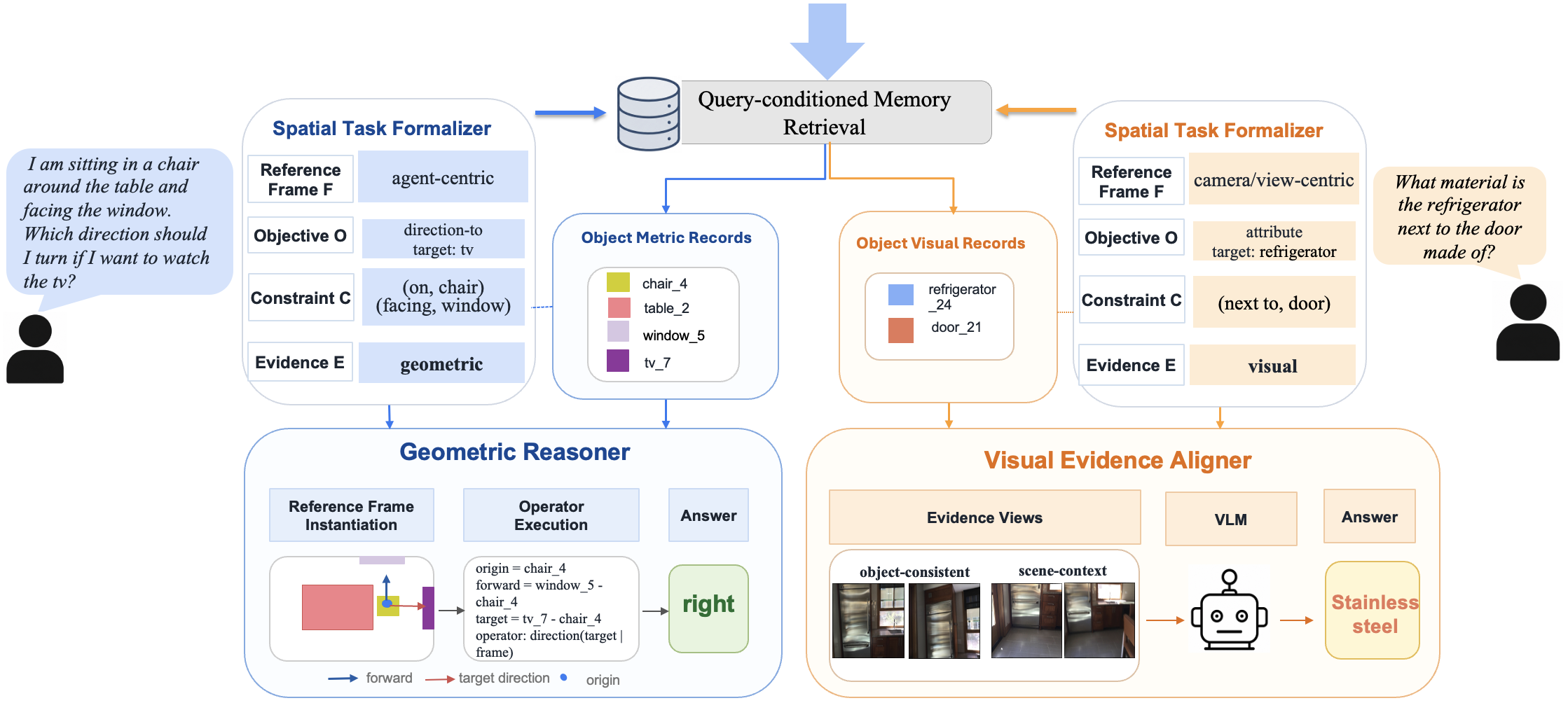}
    \caption{Overview of the BEV3SR framework.}
    \vspace{-2mm}
    \label{fig:framework}
\end{figure*}

We present the overall framework of BEV3SR in Fig.~\ref{fig:framework}. BEV3SR consists of four main components. First, we construct a BEV-indexed metric scene memory from RGB-D video observations and camera parameters (Sec.~3.1). Second, the Spatial Task Formalizer converts each natural-language query into an evidence-aware spatial task (Sec.~3.2). Based on the task specification, geometry-solvable queries are answered by the Geometric Reasoner (Sec.~3.3), while perception-dependent queries are handled by the Visual Evidence Aligner (Sec.~3.4).

\subsection{BEV-Indexed Metric Scene Memory}
BEV3SR first converts egocentric RGB-D observations into an object-centric metric scene memory. Given a video sequence
\begin{equation}
\mathcal{X}
=
\{(I_t,D_t,K_t,\mathbf{T}^{w\leftarrow c}_t)\}_{t=1}^{N},
\end{equation}
where \(I_t\) and \(D_t\) are the RGB frame and depth map, \(K_t\) is the camera intrinsic matrix, and \(\mathbf{T}^{w\leftarrow c}_t=(\mathbf{R}^{w\leftarrow c}_t,\mathbf{t}^{w\leftarrow c}_t)\) is the camera-to-world pose.

For each selected frame, we obtain panoptic segments using Mask2Former~\cite{cheng2022masked}. A valid pixel at \(u=(u_x,u_y)\) in segment \(k\) is back-projected and transformed into the world frame as
\begin{equation}
\mathbf{p}^{c}_{t,u}
=
D_t(u)K_t^{-1}
\begin{bmatrix}
u_x\\
u_y\\
1
\end{bmatrix},
\qquad
\mathbf{p}^{w}_{t,u}
=
\mathbf{R}^{w\leftarrow c}_t
\mathbf{p}^{c}_{t,u}
+
\mathbf{t}^{w\leftarrow c}_t .
\end{equation}
The lifted world points of segment $k$ form the point set
$P^w_{t,k}=\{\mathbf{p}^w_{t,u}\mid u\in k\}$, and
then we project $P^w_{t,k}$ onto the ground plane to obtain its BEV footprint
\begin{equation}
\Omega^{\mathrm{BEV}}_{t,k}
=
\{g\in\mathcal{B}
\mid
\exists\, \mathbf{p}\in P^{w}_{t,k},
\ \Pi_{\mathrm{BEV}}(\mathbf{p})=g
\},
\end{equation}
where $\Pi_{\mathrm{BEV}}$ rasterizes a world point onto the BEV grid $\mathcal{B}$.
Each view-instance $o_{t,k}$ is stored as
\begin{equation}
o_{t,k}
=
\big(
\ell_{t,k},
\Omega^{\mathrm{BEV}}_{t,k},
\mathbf{b}^{xz}_{t,k},
\boldsymbol{\mu}^{xz}_{t,k},
\mathbf{h}_{t,k},
t,k
\big),
\end{equation}
where \(\ell_{t,k}\) is the semantic label, \(\mathbf{b}^{xz}_{t,k}\) is the ground-plane bounding box, \(\boldsymbol{\mu}^{xz}_{t,k}\) is the ground-plane centroid, and \(\mathbf{h}_{t,k}=(h^{\min}_{t,k},h^{\max}_{t,k})\) is the vertical range.

Inspired by mask clustering~\cite{yan2024maskclustering} and density-based clustering~\cite{ester1996density}, we associate and consolidate repeated view-instances across frames using an adaptive instance graph
\(\mathcal{A}=(\mathcal{N}_{\mathrm{inst}},\mathcal{E}_{\mathrm{inst}})\), where each node is a view-instance \(o_{t,k}\). Edges connect semantically compatible instances whose BEV centroids, footprints, and height ranges indicate the same physical object. We treat each connected component $C_i \in \mathrm{CC}(\mathcal{A})$ as one persistent object and merge its member view-instances into a persistent metric object record $r_i$
\begin{equation}
r_i
=
\left(
id_i,
\ell_i,
\Omega_i^{\mathrm{BEV}},
\mathbf{b}^{xz}_i,
\boldsymbol{\mu}^{xz}_i,
\mathbf{h}_i,
\mathcal{S}_i
\right),
\end{equation}
where $id_i$ is the persistent object identity, $\ell_i$ is the consolidated label,
$\Omega_i^{\mathrm{BEV}}$ is the merged footprint,
$\mathbf{b}_i^{xz}$ and $\boldsymbol{\mu}_i^{xz}$ are the ground-plane bounding box and centroid,
$\mathbf{h}_i$ is the vertical range, and
$\mathcal{S}_i=\{(t,k)\mid o_{t,k}\in C_i\}$ is the set of source frame--segment ID pairs of its member view-instances, linking the persistent record to its supporting RGB observations for later visual evidence retrieval.

The final scene memory is $\mathcal{M}=(\mathcal{R},\mathcal{I})$, where $\mathcal{R}=\{r_i\}_{i=1}^{N_c}$ is the set of persistent metric object records and $\mathcal{I}$ is the associated RGB image store.
We further derive a BEV object index $\mathcal{B}^{\mathrm{idx}}(g)=\{id_i \mid g \in \Omega_i^{\mathrm{BEV}}\}$, which maps each BEV grid cell $g$ to the persistent object identities whose footprints occupy that cell.
The scene memory can then be queried at $g$ as
\begin{equation}
\operatorname{Lookup}_{\mathrm{BEV}}(g;\mathcal{M})
=
\left\{
\left(r_i,\mathcal{I}_{\mathcal{S}_i}\right)
\mid
id_i\in\mathcal{B}^{\mathrm{idx}}(g)
\right\},
\end{equation}
where $\mathcal{I}_{\mathcal{S}_i}$ denotes the RGB observations indexed by $\mathcal{S}_i$.
Thus, BEV serves as a spatial access structure that links query locations or neighborhoods to persistent metric object records in $\mathcal{R}$ and their supporting RGB evidence in $\mathcal{I}$, rather than as a top-down image.

\subsection{Spatial Task Formalizer}
\label{sec:spatial-task-formalizer}
The Spatial Task Formalizer converts each query into an evidence-aware spatial task specification before any geometric computation or visual prompting is performed.

Formally, given a question $q$, the formalizer produces
\begin{equation}
\mathcal{P}_q = F_{\mathrm{STF}}(q)
=
\left(
\mathcal{F}_q,
\mathcal{O}_q,
\mathcal{C}_q,
\mathcal{E}_q
\right),
\end{equation}
where $\mathcal{F}_q$ denotes the query-specified reference frame description, $\mathcal{O}_q$ specifies the objective, $\mathcal{C}_q$ contains symbolic spatial constraints, and $\mathcal{E}_q$ specifies required evidence type.

\paragraph{Reference-frame specification.}
The central field in $\mathcal{P}_q$ is the reference-frame description $\mathcal{F}_q$, since many 3D scene questions cannot be answered directly in any observed camera frame. We consider four common query-frame types. (1) Agent-centric frames are used when the query is defined relative to the user's situated perspective, such as \textit{my left}, or \textit{behind me}. (2) Camera/view-centric frames are used when a relation is tied to an observed or retrieved support view. (3) Object-centric frames are used only when the query provides an intrinsic object orientation or an action that defines one. (4) Scene-level allocentric frames are used when the relation is defined with respect to the global room or BEV layout, such as \textit{the top-left corner of the room}. 

The reference frame is represented by symbolic cues:
\begin{equation}
\mathcal{F}_q =
\left(
\tau_q,
\mathcal{C}^{\mathrm{frame}}_q
\right),
\end{equation}
where
\begin{equation*}
\tau_q \in
\left\{
\mathrm{Agent},
\mathrm{View},
\mathrm{Object},
\mathrm{Scene}
\right\}
\end{equation*}
denotes the frame type, and $\mathcal{C}^{\mathrm{frame}}_q$ contains the frame-defining constraints used to instantiate the coordinate  frame. These constraints are extracted from the query description.

For agent-centric frames, $\mathcal{C}^{\mathrm{frame}}_q$ includes two kinds of cues: standpoint cues that define the frame origin and heading cues that constrain the forward axis. Standpoint cues describe where the agent is situated, such as \textit{sitting on a chair} or \textit{leaning on a counter}. Heading cues describe how the agent is oriented, such as \textit{facing a table}, \textit{a bicycle on my left}, or \textit{a TV behind me}. For example, given the situation ``I am leaning on the kitchen counter and the bicycle is on my left,'' the formalizer produces
\begin{equation}
\mathcal{F}_q =
\left(
\mathrm{Agent},
\left\{
\begin{array}{l}
(\mathrm{standpoint}, \textit{kitchen counter}),\\
(\mathrm{left}, \textit{bicycle})
\end{array}
\right\}
\right).
\end{equation}
For view-centric frames, $\mathcal{C}^{\mathrm{frame}}_q$ specifies an observed or retrieved support view whose camera frame is used to interpret view-dependent relations. For example, in ``How many chairs are to the left of the table?'', the support view defines the left-right axis, while the relation \textit{left of the table} remains a task constraint over chair candidates. For object-centric frames, $\mathcal{C}^{\mathrm{frame}}_q$ identifies the oriented object or action cue that defines the object's local frame, such as \textit{the front-facing side of a microwave}. We only use this frame type when the object orientation is reliable or explicitly implied by the query. For scene-level frames, $\mathcal{C}^{\mathrm{frame}}_q$ specifies the canonical BEV or room-level coordinate convention.
\paragraph{Objective.}
The formalizer also extracts the query objective $\mathcal{O}_q$, which specifies the task type, the target description, and the query-specified relational constraints expressed in the question. 
We represent it as
\begin{equation}
\mathcal{O}_q =
\left(
\omega_q,\,
d^{\mathrm{tar}}_q,\,
\mathcal{C}^{\mathrm{rel}}_q
\right),
\end{equation}
where $\omega_q$ denotes the task type, including counting, distance computing, size measurement, direction calculation, attribute recognition, object identification, and various spatial relation; 
$d^{\mathrm{tar}}_q$ denotes the target description, such as an object category or referring phrase; 
and $\mathcal{C}^{\mathrm{rel}}_q$ denotes the query-specified relational constraints, i.e., task-level constraints expressed by spatial predicates, anchors, and spatial regions.
For example, ``How many cups are on the table?'' is formalized as
\begin{equation*}
\mathcal{O}_q =
\left(
\textsc{Count},\,
\textit{cup},\,
\left\{
\left(
\textit{on},\,
\textit{table}
\right)
\right\}
\right),
\end{equation*}

\noindent\textbf{Evidence requirement.} 
We assign each task an evidence requirement 
\begin{equation} \mathcal{E}_q \in \left\{ \mathrm{Metric}, \mathrm{Visual} \right\}. 
\end{equation} 
\begin{equation}
\begin{aligned}
\mathcal{E}_q
=&
G_{\mathrm{frame}}\left(\mathcal{F}_q\right)
\land
G_{\mathrm{obj}}\left(\mathcal{O}_q\right)
\land
G_{\mathrm{con}}\left(\mathcal{C}_q\right).
\end{aligned}
\end{equation}
Here, $G_{\mathrm{frame}}$ checks the reference frame type and whether the query reference frame can be specified symbolically, $G_{\mathrm{obj}}$ checks the task type and whether the objective can be solved by metric operations, and $G_{\mathrm{con}}$ checks whether the query constraints can be evaluated using geometric relations rather than RGB appearance. If all three conditions hold, the task is treated as metric-resolvable and passed to the Geometric Reasoner for deterministic computation. Otherwise, the task is treated as visual-required and passed to the Visual Evidence Aligner for RGB evidence retrieval.

\subsection{Geometric Reasoner}
\label{sec:geometric_reasoner}
\paragraph{Reference-frame instantiation.}
The Geometric Reasoner first instantiates the query reference frame specified by
\(\mathcal{F}_q\). Scene-level frames directly inherit the global scene coordinates.
For agent-centric frames, the agent pose is not observed in the video, so we infer a
local frame from the frame-defining constraints
\(\mathcal{C}^{\mathrm{frame}}_q\).
We search over candidate standpoint anchors and heading hypotheses,
and select the pair that best satisfies the spatial cues in the query.

We represent each frame cue as a pair \((\ell,d)\), where \(d\) is an object
description and \(\ell\) indicates its role, such as \(\mathrm{standpoint}\),
\(\mathrm{front}\), \(\mathrm{behind}\), \(\mathrm{left}\), \(\mathrm{right}\),
\(\mathrm{on}\), or \(\mathrm{next\text{-}to}\). The standpoint cue determines
candidate frame origins, directional cues constrain the agent heading, and other
relational cues help disambiguate object binding.
Given a standpoint description \(d_a\), we retrieve candidate anchor records
\(\mathcal{A}_q(d_a)\) from the memory. Each candidate anchor \(r_a \in \mathcal{A}_q(d_a)\) defines a
candidate origin \(\mathbf{o}_a \in \mathbb{R}^2\). Conditioned on \(r_a\), we retrieve local
candidate records \(\mathcal{S}_q(d;r_a)\) matching each remaining cue description
\(d\). For each candidate cue record \(r_i \in \mathcal{S}_q(d;r_a)\),
\(\mathbf{c}_i \in \mathbb{R}^2\) denotes its BEV center.

Following qualitative spatial reasoning \cite{frank1996qualitative}, we avoid
unconstrained heading regression and instead enumerate a finite set of heading
hypotheses \(\Theta(r_a)\), initialized from dominant wall normals in the BEV layout
and optionally augmented by query-induced directions.
 Each heading
\(\mathbf{h}\in\Theta(r_a)\) induces a right axis \(R_{\mathrm{cw}}\mathbf{h}\),
where \(R_{\mathrm{cw}}\) denotes a clockwise \(90^\circ\) rotation in the BEV
plane. For a candidate cue record \(r_i\), let
\(\Delta_i=\mathbf{c}_i-\mathbf{o}_a\). The signed directional score for a directional relation
\(\rho\) is
\begin{equation}
g_\rho(r_i;\mathbf{o}_a,\mathbf{h})=
\begin{cases}
\Delta_i^\top\mathbf{h}, & \rho=\textit{front},\\
-\Delta_i^\top\mathbf{h}, & \rho=\textit{behind},\\
\Delta_i^\top R_{\mathrm{cw}}\mathbf{h}, & \rho=\textit{right},\\
-\Delta_i^\top R_{\mathrm{cw}}\mathbf{h}, & \rho=\textit{left}.
\end{cases}
\end{equation}
The anchor-heading pair is selected by maximizing the number of satisfied
directional frame cues:
\begin{equation}
\begin{aligned}
\mathcal{J}(r_a,\mathbf{h})
=
&\sum_{(\rho,d)\in\mathcal{C}^{\mathrm{dir}}_q}
\max_{r_i\in\mathcal{S}_q(d;r_a)}
\mathbb{I}\!\left[
g_\rho(r_i;\mathbf{o}_a,\mathbf{h})> 0
\right], \\
(r_a^\star,\mathbf{h}_q^\star)
=
&\arg\max_{\substack{
r_a\in\mathcal{A}_q(d_a)\\
\mathbf{h}\in\Theta(r_a)}}
\mathcal{J}(r_a,\mathbf{h}).
\end{aligned}
\end{equation}
where \(\mathcal{C}^{\mathrm{dir}}_q\subseteq\mathcal{C}^{\mathrm{frame}}_q\)
contains directional frame cues. The agent-centric frame is then instantiated as
\begin{equation}
\Phi_q
=
(\mathbf{o}_q,\mathbf{h}_q,\mathbf{r}_q)
=
(\mathbf{o}_{a^\star},\mathbf{h}_q^\star,
R_{\mathrm{cw}}\mathbf{h}_q^\star).
\end{equation}

\noindent\textbf{Geometric operator execution.}
After instantiating the query frame \(\Phi_q\), the reasoner executes the
objective
\(\mathcal{O}_q=(\omega_q,d_q^{\mathrm{tar}},\mathcal{C}^{\mathrm{rel}}_q)\)
over the metric object records. We first retrieve candidate records matching
\(d_q^{\mathrm{tar}}\), and then retain those satisfying all relational constraints:
\begin{equation}
\mathcal{V}_q =
\left\{
r_i\in\mathcal{R}_q(d_q^{\mathrm{tar}})
\;\middle|\;
\psi_c(r_i;\Phi_q)=1,\ 
\forall c\in\mathcal{C}^{\mathrm{rel}}_q
\right\},
\end{equation}
where \(\mathcal{R}_q(d)\subseteq\mathcal{R}\) denotes records matched to
description \(d\), and \(\psi_c\) tests whether \(r_i\) satisfies constraint \(c\)
in the query frame \(\Phi_q\).

The answer is then computed by the task-specific operator:
\begin{equation}
a_q =
\operatorname{Exec}_{\omega_q}(\mathcal{V}_q,\Phi_q).
\end{equation}
For example, counting returns \(|\mathcal{V}_q|\), direction maps signed
projections to the required answer space, distance and size use metric object
centers and bounding boxes, and vertical or support relations use height ranges
and BEV footprint overlap. If the query cannot be resolved geometrically, it is
passed to the Visual Evidence Aligner.

\subsection{Visual Evidence Aligner}
\label{sec:visual_evidence_aligner}
For queries requiring visual evidence, the Visual Evidence Aligner reuses the
same reference-frame instantiation, when needed, and query-to-object binding
over the BEV-indexed memory to identify an evidence record set
\(\mathcal{R}^{\mathrm{ev}}_q\). This set includes records matched to the target
description and constraint-related records that localize the relevant
visual region. Each evidence record \(r_i\in\mathcal{R}^{\mathrm{ev}}_q\) is linked
to a set of supporting images \(\mathcal{I}^{\mathrm{sup}}(r_i)\subseteq\mathcal{I}\).

The aligner selects images under a total budget \(K\) from two complementary sources. Object-consistent images are selected from the supporting images of \(\mathcal{R}^{\mathrm{ev}}_q\), providing clear observations of the evidence records. Scene-context images provide complementary context and can come from surrounding regions in different directions around the evidence records, or from globally diverse scene images when broader scene understanding is needed.
Given the current selected image set \(\mathcal{I}^{\mathrm{vis}}_q\), the next
image is chosen greedily:
\begin{equation}
I^{\star}
=
\arg\max_{I_f\notin\mathcal{I}^{\mathrm{vis}}_q}
\left[
\lambda^{\omega_q}_{o}\Delta_{\mathrm{obj}}(I_f)
+
\lambda^{\omega_q}_{c}\Delta_{\mathrm{ctx}}(I_f)
\right],
\end{equation}
where \(\Delta_{\mathrm{obj}}\) measures additional evidence-record coverage and
\(\Delta_{\mathrm{ctx}}\) measures additional scene-context coverage. The weights
\(\lambda^{\omega_q}_{o}\) and \(\lambda^{\omega_q}_{c}\) are determined by the task
type \(\omega_q\). We use a balanced weighting by default, while attribute queries place greater weight on object-consistent evidence. The
final selected set satisfies
\(|\mathcal{I}^{\mathrm{vis}}_q|\le K\),
and is provided to the VLM for answer generation.

\section{Experiments}
\label{sec:experiments}

\subsection{Experimental Setup}
We evaluate BEV3SR on three 3D spatial reasoning benchmarks: ScanQA~\cite{azuma2022scanqa}, SQA3D~\cite{ma2022sqa3d}, and VSI-Bench~\cite{yang2025thinking}. ScanQA focuses on object-centric 3D scene QA, while SQA3D introduces situation descriptions and agent-centric reasoning. We report results on ScanQA val and SQA3D test, using EM, BLEU-4, METEOR, ROUGE, and CIDEr for ScanQA, and EM/EM-R~\cite{huang2024embodied} for SQA3D. For VSI-Bench, we follow the static spatial evaluation setting and exclude route planning and appearance order. We therefore report Avg.-6 over the remaining six static spatial categories: object counting, absolute distance, object size, room size, relative distance, and relative direction.
We use Qwen3-VL-8B as the default backbone VLM, and additionally evaluate InternVL3.5-8B and LLaVA-OneVision-1.5-8B to assess cross-model generality. 
All experiments are conducted on a single NVIDIA A6000 GPU with 48GB memory.
\subsection{Results}
\subsubsection{Generalization Across VLMs.}
\cref{tab:uniform_vs_bevtriage_multimodel} shows that BEV3SR consistently improves over the baseline across different VLM backbones under the same $K{=}16$ frame budget, where the baseline denotes uniform 16-frame prompting with the corresponding VLM backbone.
On SQA3D, BEV3SR improves EM by +11.0, +8.3, and +11.1 for Qwen3-VL-8B, InternVL3.5-8B, and LLaVA-OneVision-1.5-8B, respectively, with similar gains in EM-R. 
On ScanQA, BEV3SR also improves EM across all backbones, although captioning-style metrics such as CIDEr and METEOR are more sensitive to answer style and paraphrasing. 
These results suggest that the gains come from the proposed evidence-aware framework and modular geometric reasoning / visual evidence alignment, rather than model-specific behavior.

\subsubsection{Comparison to Prior Work.}
\paragraph{Comparison with 3D-aware Models.}
We compare BEV3SR with representative task-specific models, 3D LLM-based models, and recent vision-LLM approaches in \cref{tab:scanqa_sqa3d_val}.Among the recent related approaches, GPT4Scene \cite{GPT4Scene} uses BEV-aligned object markers, Video-3D LLM \cite{zheng2024video3dllmlearningpositionaware} incorporates 3D positional information, and LLaVA-3D \cite{zhu2025llava} constructs 3D-aware tokens. Unlike these approaches, which primarily expose or encode 3D information for VLM reasoning, BEV3SR treats 3D structure as a queryable metric memory for explicit query-conditioned reasoning. This enables query-specific reference-frame instantiation, deterministic geometric computation over grounded objects, and constraint-aware visual evidence retrieval under a fixed frame budget, rather than leaving these operations implicit within VLM inference.

BEV3SR achieves the best performance on SQA3D among the compared methods, reaching 62.6 EM and 64.8 EM-R. On ScanQA, it obtains 30.6 EM matching LLaVA-3D. These results indicate that BEV3SR is particularly effective for situated 3D reasoning while remaining competitive on general 3D scene QA without task-specific training. As shown in \cref{tab:vsi_static6}, BEV3SR further improves the baseline from 53.8 to 63.8 Avg.-6 on the static spatial subset of VSI-Bench, a 10.0-point gain, and achieves the best overall score among the compared methods. The largest gains occur in absolute distance, relative distance, and relative direction, further supporting the benefit of explicit metric and reference-frame-aware reasoning for spatial tasks.

\begin{table}[t]
  \centering
  \scriptsize
  \setlength{\tabcolsep}{2.2pt}
  \renewcommand{\arraystretch}{1.05}
   \begin{threeparttable}
  \caption{Baseline vs.\ BEV3SR across different VLMs. We report EM (exact match), EM-R (refined exact match), BLEU-4 (B-4), METEOR (M), ROUGE (R), and CIDEr (C).}
\vspace{-2mm}
  \label{tab:uniform_vs_bevtriage_multimodel}
  \begin{tabular}{@{}L{0.29\columnwidth}lcc|ccccc@{}}
    \toprule
    \multirow{2}{*}{\textbf{VLM}} & \multirow{2}{*}{\textbf{Method}} &
    \multicolumn{2}{c}{\textbf{SQA3D (test)}} &
    \multicolumn{5}{c}{\textbf{ScanQA (val)}} \\
    \cmidrule(lr){3-4}\cmidrule(lr){5-9}
    & & EM & EM-R & EM & B-4 & M & R & C \\
    \midrule

    \multirow{2}{*}{Qwen3-VL-8B~\cite{bai2025qwen3}}          & Baseline     & 51.6 & 55.0 & 27.8 & 15.9 & 15.8 & 41.9 & 82.6 \\
                         & BEV3SR  & 62.6 & 64.8 & 30.6 & 16.0 & 18.5 & 46.4 & 95.6 \\
    \midrule

    \multirow{2}{*}{InternVL3.5-8B~\cite{wang2025internvl3}}       & Baseline    & 52.9 & 55.2 & 28.2 & 14.2 & 17.4 & 44.5 & 90.4 \\
                         & BEV3SR  & 61.2 & 62.5 & 30.2   & 14.7   & 16.9   & 45.4   & 90.7   \\
    \midrule

    \multirow{2}{*}{\parbox{2cm}{LLaVA-OneVision-1.5-8B~\cite{LLaVA-OneVision-1.5}}} & Baseline     & 49.2 & 50.1 & 22.1 & 4.2  & 13.3 & 34.8 & 66.0 \\
                           & BEV3SR & 60.3 & 61.4 & 25.4 & 5.7  & 13.3 & 36.7 & 69.1   \\
    \bottomrule
  \end{tabular}
    \begin{tablenotes}[flushleft]
    \footnotesize
    \item \emph{Note:} The RGB-only prompting baselines are used as controlled VLM
    references under the same backbone and frame budget, rather than as equal input
    modality baselines.
  \end{tablenotes}
  \end{threeparttable}
\vspace{-2mm}
\end{table}

\begin{table}[tb]
  \centering
  \caption{Comparison with SOTA models on SQA3D and ScanQA. (PC: point cloud,  Intr./Pose: camera intrinsics and poses.)}
  \vspace{-2mm}
  \label{tab:scanqa_sqa3d_val}
  \footnotesize
  \setlength{\tabcolsep}{3.2pt}
  \renewcommand{\arraystretch}{1.08}

  \begin{adjustbox}{max width=\linewidth}
  \begin{tabular}{l c cc|ccccc}
    \toprule
    \multirow{2}{*}{\textbf{Methods}} &
    \multirow{2}{*}{\textbf{Scene Input}} &
    \multicolumn{2}{c|}{\textbf{SQA3D (test)}} &
    \multicolumn{5}{c}{\textbf{ScanQA (val)}} \\
    \cmidrule(lr){3-4}\cmidrule(lr){5-9}
    & & EM & EM-R & EM & B-4 & M & R & C \\
    \midrule

    \multicolumn{9}{l}{\textit{Task-Specific Model}} \\
    ScanQA~\cite{azuma2022scanqa}
      & PC+RGB & -- & -- & 21.1 & 10.1 & 13.1 & 33.3 & 64.9 \\
    SQA3D~\cite{ma2022sqa3d}
      & PC+RGB & 46.6 & -- & -- & -- & -- & -- & -- \\
    3D-VisTA~\cite{3dvista}
      & PC+RGB & 48.5 & -- & 22.4 & 10.4 & 13.9 & 35.7 & 69.6 \\
    \midrule

    \multicolumn{9}{l}{\textit{3D LLM Based Model}} \\
    3D-LLM~\cite{3dllm}
      & PC & -- & -- & 20.5 & 12.0 & 14.5 & 35.7 & 69.4 \\
    LL3DA~\cite{chen2024ll3da}
      & PC & -- & -- & -- & 13.5 & 15.9 & 37.3 & 76.8 \\
    PQ3D~\cite{zhu2024unifying}
      & PC+RGB & 47.1 & -- & -- & -- & -- & -- & -- \\
    Scene-LLM~\cite{fu2024scenellmextendinglanguagemodel}
      & PC+RGB & 53.6 & -- & 27.2 & 11.7 & 15.8 & 35.9 & 80.0 \\
    Grounded3D-LLM~\cite{chen2024grounded3dllm}
      & PC & -- & -- & -- & 13.4 & -- & -- & 72.7 \\
    LEO~\cite{huang2024embodied}
      & PC & 50.0 & 52.9 & 24.5 & 13.2 & 20.0 & 49.2 & 101.4 \\
    Chat-Scene~\cite{huang2024chat}
      & PC+RGB & 54.6 & 57.5 & 21.6 & 14.3 & 18.0 & 41.6 & 87.7 \\
    LLaVA-3D~\cite{zhu2025llava}
      & RGB-D+Intr./Pose & 60.1 & -- & \textbf{30.6} & \textbf{16.4}
      & \textbf{20.8} & \textbf{49.6} & \textbf{103.1} \\
    \midrule

    \multicolumn{9}{l}{\textit{Vision LLM Based Model}} \\
    GPT4Scene~\cite{GPT4Scene}
      & RGB+Intr./Pose & 59.4 & 62.4 & 28.2 & 15.5 & 18.9 & 46.5 & 96.3 \\
    Video-3DLLM~\cite{zheng2024video3dllmlearningpositionaware}
      & RGB-D+Intr./Pose & 58.6 & -- & 30.1 & 16.2 & 19.8 & 49.0 & 102.1 \\
    \textbf{BEV3SR}
      & RGB-D+Intr./Pose & \textbf{62.6} & \textbf{64.8}
      & \textbf{30.6} & 16.0 & 18.5 & 46.4 & 95.6 \\
    \bottomrule
  \end{tabular}
  \end{adjustbox}

  \vspace{-4mm}
\end{table}

\begin{table}[!t]
\centering
\setlength{\heavyrulewidth}{0.05em}
\setlength{\lightrulewidth}{0.03em}
\setlength{\cmidrulewidth}{0.03em}
\caption{Comparison on the static spatial subset of VSI-Bench.}
\vspace{-2mm}
\label{tab:vsi_static6}
\tiny
\setlength{\tabcolsep}{3.0pt}
\renewcommand{\arraystretch}{1.02}
\resizebox{\linewidth}{!}{%
\begin{tabular}{>{\raggedright\arraybackslash}p{0.30\linewidth}ccccccc}
\toprule
\textbf{Model}
& \textbf{Avg.}
& \multicolumn{4}{c}{\textbf{Num.}}
& \multicolumn{2}{c}{\textbf{MC}} \\
\cmidrule(lr){3-6}\cmidrule(lr){7-8}
&
& \shortstack{Obj.\\Cnt.}
& \shortstack{Abs.\\Dist.}
& \shortstack{Obj.\\Size}
& \shortstack{Room\\Size}
& \shortstack{Rel.\\Dist.}
& \shortstack{Rel.\\Dir.} \\
\midrule
\multicolumn{8}{l}{\textit{2D MLLM}} \\
SpatialMLLM-4B~\cite{wu2026spatial}
& 49.3 & 65.3 & 34.8 & 63.1 & 45.1 & 41.3 & 46.2 \\
VG-LLM-8B~\cite{zheng2026learning}
& 52.3 & 67.9 & 37.7 & 58.6 & \textbf{62.0} & 46.6 & 40.7 \\

\midrule
\multicolumn{8}{l}{\textit{Open source}} \\
InternVL3.5-8B~\cite{wang2025internvl3}
& 57.3 & \textbf{70.2} & 42.7 & \textbf{67.4} & 60.9 & 54.6 & 48.1 \\
LLaVA-OneVision-1.5-8B~\cite{LLaVA-OneVision-1.5}
& 36.2 & 44.3 & 9.5 & 45.2 & 42.4 & 37.7 & 38.1 \\

\midrule
Qwen3-VL-8B~\cite{bai2025qwen3} (baseline)
& 53.8 & 61.3 & 40.2 & 69.9 & 57.5 & 46.1 & 47.6 \\
\textbf{BEV3SR}
& \textbf{63.8} & 67.9 & \textbf{57.2} & 52.5
& 60.8 & \textbf{60.7} & \textbf{83.5} \\
\bottomrule
\end{tabular}
}
\vspace{-1.0em}
\end{table}
\noindent\textbf{Contextual training-free comparison.}
We include these training-free methods in \cref{tab:training_free_context} as contextual references rather than strict equal-setting baselines, since they use different backbone scales and inference protocols. SpatialMind \cite{zhang2026spatial} relies on a much larger Qwen2.5-VL-72B backbone, while CoV \cite{zhao2026cov} performs API-based active multi-step view search. In contrast, BEV3SR uses an Qwen3-VL-8B backbone with a fixed 16-frame visual budget and achieves strong performance on situated egocentric reasoning.
\begin{table}[t]
\centering
\caption{Contextual comparison with training-free methods.}
\vspace{-2mm}
\label{tab:training_free_context}
\scriptsize
\setlength{\tabcolsep}{3.5pt}
\begin{tabularx}{\linewidth}{l>{\raggedright\arraybackslash}Xcc}
\toprule
Method & Backbone & ScanQA & SQA3D \\
\midrule
SpatialMind~\cite{zhang2026spatial}
& Qwen2.5-VL-72B~\cite{bai2025qwen25vltechnicalreport}
& B-1 38.0 & EM 39.2 \\

CoV~\cite{zhao2026cov}
& GPT-4.1~\cite{achiam2023gpt} / Gemini~\cite{comanici2025gemini}
& EM 31.9 & EM 51.1 \\

BEV3SR
& Qwen3-VL-8B~\cite{bai2025qwen3}
& EM 30.6 / B-1 41.0 & EM 62.6 \\
\bottomrule
\end{tabularx}
\vspace{-1.0em}
\end{table}

\subsection{Ablation Study}
\subsubsection{Effectiveness of Geometric Reasoner.}
Table~\ref{tab:ablation_combined}(a) evaluates Geometric Reasoner (GR) on the geometry-solvable subset of SQA3D. Compared with baseline, GR improves EM/EM-R from 43.2/44.0 to 65.5/65.9, yielding +22.3/+21.9 gains. Baseline denotes 16-frame uniformly sampled RGB prompting with Qwen3VL-8B. We also compare with BEV, that provides relevant object instance IDs and labels to the VLM without explicit geometric computation. This cue-injection baseline does not consistently improve performance and even degrades EM from 43.2 to 41.7, indicating that geometry-intensive queries cannot be reliably solved by additional BEV text hints alone. Overall, the ablation confirms that explicit geometric computation in GR over BEV-indexed metric records is crucial for query-specified spatial reasoning. 

\begin{table}[t]
\centering
\caption{Ablation studies on the SQA3D validation set: 
(a) effectiveness of GR on the geometry-solvable subset; 
(b) end-to-end contribution of GR and VEA on the full set.}
\label{tab:ablation_combined}
\vspace{-2mm}
\scriptsize
\begin{minipage}[t]{0.35\columnwidth}
\centering
{\footnotesize  (a) GR effectiveness}
\vspace{0.5mm}

\scriptsize
\setlength{\tabcolsep}{1.5pt}
\renewcommand{\arraystretch}{0.95}
\begin{tabular}{lcccc}
\toprule
\textbf{Method}
& \textbf{EM} & \textbf{$\Delta$}
& \textbf{EM-R} & \textbf{$\Delta$} \\
\midrule
Baseline
& 43.2 & -- & 44.0 & -- \\
+ BEV
& 41.7 & $-1.5$ & 42.3 & $-1.7$ \\
+ GR
& \textbf{65.5} & \textbf{+22.3}
& \textbf{65.9} & \textbf{+21.9} \\
\bottomrule
\end{tabular}
\end{minipage}
\hfill
\begin{minipage}[t]{0.48\columnwidth}
\centering
{\footnotesize (b) Component contribution}
\vspace{0.5mm}

\scriptsize
\setlength{\tabcolsep}{1.0pt}
\renewcommand{\arraystretch}{0.95}
\begin{tabular}{lcccc}
\toprule
\textbf{Method}
& \textbf{EM} & \textbf{$\Delta$}
& \textbf{EM-R} & \textbf{$\Delta$} \\
\midrule
Baseline
& 51.0 & -- & 53.6 & -- \\
+ GR
& 59.4 & +8.4 & 61.8 & +8.2 \\
+ VEA
& 52.9 & +1.9 & 55.3 & +1.7 \\
+ GR + VEA
& \textbf{61.3} & \textbf{+10.3}
& \textbf{63.5} & \textbf{+9.9} \\
\bottomrule
\end{tabular}
\end{minipage}
\vspace{-6mm}
\end{table}
\subsubsection{Effectiveness of Visual Evidence Aligner.}
We ablate the Visual Evidence Aligner (VEA) by comparing it against baseline under the same $K{=}16$ frame budget. In \Cref{tab:scanqa_ablation_uniform_keyframe}, BEV3SR-VEA improves EM from 27.8 to 30.6 (+2.8) and CIDEr from 82.6 to 95.6 (+13.0) on ScanQA (val), indicating stronger evidence grounding for open-vocabulary 3D QA.On perception-dependent SQA3D (val) questions, VEA similarly improves EM/EM-R from 55.7/59.3 to 58.7/62.1 (+3.0/+2.8), showing consistent gains in visual evidence grounding.

\begin{table}[t]
\centering
\caption{Ablation study for effectiveness of Visual Evidence Aligner (VEA) on ScanQA validation set.}
\vspace{-2mm}
\label{tab:scanqa_ablation_uniform_keyframe}
\scriptsize
\setlength{\tabcolsep}{3.5pt}
\begin{tabular}{lcccccccc}
\toprule
\textbf{Method} & \textbf{EM} & \textbf{B-1} & \textbf{B-2} & \textbf{B-3} & \textbf{B-4} & \textbf{M} & \textbf{R} & \textbf{C} \\
\midrule
Baseline
& 27.8 & 35.4 & 23.2 & 18.2 & 15.9 & 15.8 & 41.9 & 82.6 \\
VEA 
& 30.6 & 41.0 & 27.0 & 20.7 & 16.0 & 18.5 & 46.4 & 95.6 \\
\midrule
$\Delta$ 
& \textbf{+2.8} & \textbf{+5.6} & \textbf{+3.8} & \textbf{+2.5} & \textbf{+0.1} & \textbf{+2.7} & \textbf{+4.5} & \textbf{+13.0}\\
\bottomrule
\end{tabular}%
\vspace{-2.5mm}
\end{table}
\subsubsection{End-to-End Contribution on Full Validation Set}
Since STF provides the common task specification required by both
downstream pathways, we isolate the end-to-end contributions of GR and VEA while keeping STF fixed.
To compare both pathways under the same evaluation setting, we
aggregate their pathway-specific results over the full 3,261 question
SQA3D validation set. In the GR-only setting, geometric solvable questions
use GR while perception-dependent questions retain the baseline
predictions, the VEA-only setting is defined analogously.
As shown in Tab.~\ref{tab:ablation_combined}(b), GR alone improves overall
EM/EM-R from 51.0/53.6 to 59.4/61.8. Combining both pathways reaches 61.3/63.5.

\subsection{Robustness to RGB-Estimated Geometry}
\label{sec:rgb_robustness}
To assess BEV3SR's dependence on sensor-provided geometry, we construct an RGB-only variant that jointly estimates depth, camera intrinsics, and camera poses from RGB using MapAnything~\cite{keetha2026mapanything}, while keeping the downstream pipeline unchanged. The estimated geometry is substantially imperfect, with an average depth AbsRel of 20.6\%, focal-length error of 4.7\%, and camera-pose ATE of 0.34\,m. Despite these errors, Table~\ref{tab:rgb_robustness} shows that BEV3SR (RGB-only) still improves the RGB baseline from 51.0 to 56.5 EM on SQA3D validation and from 27.8 to 29.0 on ScanQA validation. Sensor-provided geometry further increases performance to 61.3 and 30.6, respectively. These results show that BEV3SR remains effective with imperfect RGB-estimated geometry rather than requiring pristine geometric inputs, while also quantifying the benefit of more accurate depth and camera parameters.
\begin{table}[!t]
\centering

\begin{minipage}[t]{0.50\columnwidth}
\centering
\caption{Robustness of BEV3SR to RGB-estimated geometry. All results are EM (\%).}
\vspace{-2mm}
\label{tab:rgb_robustness}
\footnotesize
\setlength{\tabcolsep}{2.5pt}
\renewcommand{\arraystretch}{0.98}
\begin{tabular}{@{}lcc@{}}
\toprule
Method & SQA3D & ScanQA \\
\midrule
Baseline          & 51.0 & 27.8 \\
BEV3SR (RGB-only) & 56.5 & 29.0   \\
BEV3SR            & 61.3 & 30.6 \\
\bottomrule
\end{tabular}
\end{minipage}
\hfill
\begin{minipage}[t]{0.40\columnwidth}
\centering
\caption{Accuracy of the Spatial Task Formalizer (STF) on SQA3D (val).}
\vspace{-2mm}
\label{tab:stf_diagnostic}
\footnotesize
\setlength{\tabcolsep}{3pt}
\renewcommand{\arraystretch}{0.98}
\begin{tabular}{@{}lc@{}}
\toprule
STF Field & Acc. (\%) \\
\midrule
Reference frame      & 95.8 \\
Objective             & 91.6 \\
Query constraint      & 92.2 \\
Evidence requirement  & 98.4 \\
\bottomrule
\end{tabular}
\end{minipage}

\vspace{-1em}
\end{table}
\subsection{Error Diagnostics}
\paragraph{Spatial Task Formalizer.}
For diagnostic purposes, we evaluate reference-frame parsing based on the frame type and frame-defining constraints, objective parsing based on the task type and target description, query-specified relational constraints, and evidence-requirement prediction on SQA3D validation dataset. As shown in Table~\ref{tab:stf_diagnostic}, STF achieves high accuracy across all four fields. The results indicate that STF reliably converts queries into structured task specifications.

\noindent\textbf{Reference-Frame Instantiation.}
We analyze the robustness of reference-frame instantiation on 1,120
geometry-solvable SQA3D validation questions. We compare the estimated agent
pose (standpoint and heading) with the ground-truth situation pose, using
$1\,\mathrm{m}$ and $30^\circ$ as the position and heading error thresholds,
respectively. As shown in \cref{tab:pose_error_bins}, QA accuracy is highest
when both the standpoint and heading are accurate, and decreases when either
component is erroneous. This shows that geometric reasoning is sensitive to
reference-frame quality. Position errors can affect local object retrieval,
while heading errors can alter front/behind/left/right relations.
\begin{table}[!t]
\centering
\caption{QA accuracy under different reference-frame error regimes on SQA3D val.}
\vspace{-2mm}
\label{tab:pose_error_bins}
\scriptsize
\setlength{\tabcolsep}{3pt}
\renewcommand{\arraystretch}{1.02}
\begin{tabular}{@{}p{0.64\columnwidth}cc@{}}
\toprule
\textbf{Reference-frame Regime} & \textbf{\#Q} & \textbf{EM} \\
\midrule
Good position, good heading
($e_p \leq 1\,\mathrm{m}$, $e_\theta \leq 30^\circ$) & 547 & 76.97 \\
Good position, bad heading
($e_p \leq 1\,\mathrm{m}$, $e_\theta > 30^\circ$) & 169 & 58.58 \\
Bad position, good heading
($e_p > 1\,\mathrm{m}$, $e_\theta \leq 30^\circ$) & 142 & 65.49 \\
Bad position, bad heading
($e_p > 1\,\mathrm{m}$, $e_\theta > 30^\circ$) & 262 & 51.15 \\
\bottomrule
\end{tabular}
\vspace{-2.0em}
\end{table}

\noindent\textbf{Query-to-Object Binding.}
Using ground-truth involved-object IDs on ScanQA validation, we evaluate whether
query-relevant object references are correctly bound to persistent records in the
BEV-indexed scene memory. Query-to-object binding achieves $90.5\%$ accuracy,
indicating reliable grounding of query-relevant objects before task execution.
The retrieved evidence views generally preserve these grounded objects, while remaining failures mainly arise for objects with sparse, degraded visual
observations or incomplete BEV records.

\section{Conclusion}
In this work, we introduce BEV3SR, a training-free and backbone-agnostic framework for video-based 3D scene question answering. BEV3SR bridges the gap between fragmented egocentric video observations and query-conditioned, evidence-grounded 3D spatial reasoning. It converts RGB-D videos with camera parameters into a BEV-indexed metric scene memory, and formalizes each question as an evidence-aware spatial task. Based on  evidence, BEV3SR performs either deterministic geometric reasoning in the query-specified reference frame or BEV-guided visual evidence alignment for VLM grounding. Experiments on ScanQA, SQA3D and VSI-Bench show consistent improvements over baseline across different VLM backbones, with strong gains on spatially intensive QA. These results suggest that explicit metric scene memory, query-defined reference-frame, and modular evidence alignment provide a practical and interpretable path toward reliable 3D spatial reasoning from video.

\section*{Acknowledgement}
This work is supported in part by NSF award \# 2551891.

{
    \small
    \bibliographystyle{ieeenat_fullname}
    \bibliography{main}
}

\end{document}